\documentclass[10pt]{article}

\usepackage{arxiv}

\usepackage[utf8]{inputenc}
\usepackage[T1]{fontenc}
\usepackage{microtype}
\usepackage{hyperref}
\usepackage{url}
\usepackage{booktabs}
\usepackage{amsfonts}
\usepackage{amsmath}
\usepackage{amsthm}
\usepackage{algorithm}
\usepackage{algorithmic}
\usepackage{subcaption}
\usepackage{graphicx}
\usepackage{multirow}
\usepackage[capitalize]{cleveref}
\usepackage{bm}
\usepackage{xcolor}
\usepackage{natbib}

\newtheorem{definition}{Definition}

\definecolor{darkblue}{rgb}{0, 0, 0.5}
\hypersetup{colorlinks=true, citecolor=darkblue, linkcolor=darkblue, urlcolor=darkblue}

\title{Bucketing the Good Apples: \\A Method for Diagnosing and Improving Causal Abstraction}

\author{
  \textbf{Li Puyin}\thanks{Equal first authorship.}  $^{\ \diamondsuit}$,
  \textbf{Jiyuan Tan}\footnotemark[1]  $^{\ \diamondsuit}$,
  \textbf{Ahmad Jabbar}$^{\diamondsuit}$,
  \textbf{Thomas Icard}\thanks{Equal senior authorship.}  $^{\ \diamondsuit}$,
  \textbf{Atticus Geiger}\footnotemark[2]  $^{\ \spadesuit}$  \\[6pt]
  $^{\diamondsuit}$Stanford University \quad $^{\spadesuit}$Goodfire \\
  \texttt{\{puyinli, jiyuantan, jabbar, icard\}@stanford.edu}\\
  \texttt{atticus@goodfire.ai}\\[4pt]
}

\date{}

\begin{document}

\maketitle

\begin{abstract}
We present a method for diagnosing interpretation in neural networks by identifying an input subspace where a proposed interpretation is highly faithful. Our method is particularly useful for causal-abstraction-style interpretability, where a high-level causal hypothesis is evaluated by interchange interventions. Rather than treating interchange intervention accuracy as a single global summary, we refine this framework by partitioning the input space into well-interpreted and under-interpreted regions according to pairwise interchange-intervention behavior. This turns causal abstraction from a purely global evaluation into a more diagnostic tool: it not only measures whether an interpretation works, but also reveals where it works, where it fails, and what distinguishes the two cases. This diagnostic view also provides practical heuristics for improving interpretations. By analyzing the structure of the well-interpreted and under-interpreted regions, we can identify missing distinctions in a high-level hypothesis, discover previously unmodeled intermediate variables, and combine complementary partial interpretations into a stronger one. We instantiate this idea as a simple four-step recipe and show that it yields informative error analyses across multiple causal abstraction settings. In a toy logic task, recursively applying the recipe recovers a high-level hypothesis from scratch. More broadly, our results suggest that partitioning the input space is a useful step toward more precise, constructive, and scalable mechanistic interpretability.\footnote{We provide the code base for this paper at \url{https://github.com/Paulineli/apple-bucket}.}
\end{abstract}

\section{Introduction}

As Language Models (LMs) have scaled in complexity, the mechanistic interpretability community has developed a broad set of tools for studying the representations and algorithms underlying model behavior, including gradient-based attribution \citep{sundararajan2017axiomatic}, activation-based localization \citep{meng2022locating,meng2022mass}, and feature decomposition with Sparse Autoencoders (SAEs) \citep{bricken2023towards,cunningham2023sparse}. Among these paradigms, causal abstraction provides a particularly rigorous framework: given a task, a high-level causal model and an alignment between high-level variables and internal neural representations, it asks whether the model and the hypothesis agree under counterfactual interventions \citep{geiger2021causal, geiger2025causal}. This framework has been shown to be suitable for a wide variety of tasks \citep{wu2024pyvene, arora-etal-2024-causalgym, huanginternal, boguraev-etal-2025-causal}.

In practice, causal abstraction is usually evaluated through \emph{interchange interventions}, summarized by a single scalar metric, \emph{interchange intervention accuracy} (IIA) \citep{geiger2021causal, geiger2025causal}. The IIA score provides a simple metric to measure how well hypotheses align with neural models, but it says little about \emph{where} a hypothesis is faithful, and intermediate scores are especially hard to interpret \citep{makelov2023subspace,wu2024reply,meloux2025everything}. Moreover, although methods such as DAS make it easier to find candidate alignments, the overall workflow remains largely evaluative rather than constructive: it can tell us that a hypothesis is imperfect, but not how to improve it.

We address this gap by shifting the unit of analysis from a global score to the structure of the input space. Given a low-level model \(\mathcal L\), a high-level hypothesis \(\mathcal H\), and an alignment \(\Pi\), we partition the input space into well-interpreted \emph{target buckets} with high IIA and a complementary bucket that captures the remaining failure modes. The key question is no longer just whether an abstraction works on average, but \emph{for which inputs} it is actually faithful. This yields a more informative diagnosis and turns abstraction failures into evidence for refinement. Similar ideas of exploiting structure in observational or interventional partitions have appeared in causal feature learning \citep{chalupka2014visual,chalupka2017causal}.

Our main contribution is a practical four-step pipeline for diagnosing and improving causal abstractions: specify a reliable task and input space, obtain a candidate alignment, partition the input space by pairwise interchangeability to identify nearly interchange-consistent subsets, and train a classifier to generalize this diagnosis beyond the analyzed sample. Across fine-tuned and pretrained models of different sizes, and across alignment methods including full-vector patching, DAS, and MDAS, we show that this bucketing procedure is broadly useful for diagnosing causal abstractions. Empirically, we validate it on logic, entity binding \citep{gur2025mixing}, and factual recall \citep{geva2021transformer,hernandez2023linearity,geva2023dissecting,huang2024ravel}. In the toy logic task, recursively applying the recipe supports iterative refinement of the high-level hypothesis itself, showing how causal abstraction can move from post hoc evaluation toward constructive hypothesis discovery.

\begin{figure}[t]
    \centering
    \includegraphics[width=\linewidth]{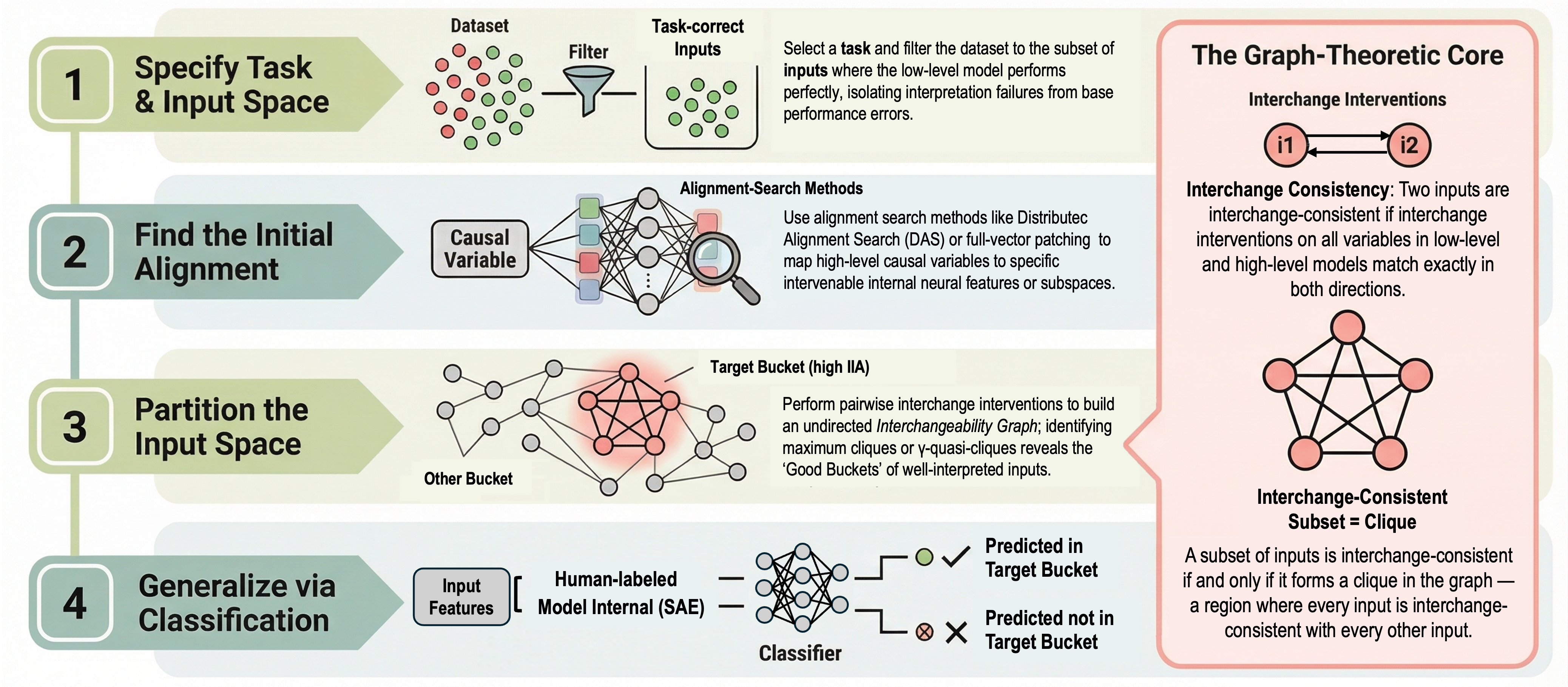}
    \caption{\textbf{Four-step interpretation diagnosis pipeline.} Given a causal abstraction for a task-performing model, we focus on task-correct inputs, identify an alignment, partition the input space by pairwise interchangeability, and train a classifier to generalize the diagnosis.}
    \label{fig:teaser}
\end{figure}

The rest of the paper is organized as follows. Section~2 reviews the causal abstraction framework and the alignment and feature-learning tools used in this work. Section~3 formalizes interchange-consistent subsets and introduces our diagnosis pipeline. Section~4 evaluates the method on three settings of increasing complexity, and Section~5 concludes with limitations and future directions.
\section{Preliminaries}


Causally abstracting an LM involves hypothesizing and verifying that a higher-level causal model is an abstraction of the lower-level LM. Causal models---familiar from the literature on causal inference (e.g., \citealt{pearl2009causality})---comprise variables, on which interventions predict counterfactual behavior. After hypothesizing alignments of the causal variables within LM representations and verifying the predicted counterfactual behavior for all such representations of an LM with appropriate interventions, the higher-level model can be taken to causally abstract the LM, if the corresponding interventions on the corresponding causal variables predict the same counterfactual output. We adopt notation from \cite{geiger2023causal}.


\textbf{Causal Abstraction.} Specifically, we use $\displaystyle \mathcal{H} ,\mathcal{L}$ to represent a high-level and a low-level causal model, respectively. For each variable $\displaystyle X$, we notate its domain as $\displaystyle \mathbb{V}_{X}$. For   $\displaystyle X$ in $\displaystyle \mathcal{H}$, $\displaystyle ( \pi_X ,\tau _{X})$ is an alignment that pairs a high-level variable $\displaystyle X$ with a set of low-level variables $\displaystyle \pi_X$ and maps variable values of $\Pi_{X}$ to variable values of $X$, i.e.,
$\displaystyle \tau _{X} :\prod _{Z\in \pi_X}\mathbb{V}_{Z}\rightarrow \mathbb{V}_{X}$.
Moreover, we define an alignment $\displaystyle \Pi $ of two models $\displaystyle \mathcal{H} ,\mathcal{L}$ as a collection of variable alignments, i.e., $\displaystyle \Pi =(\{\pi_X\} ,\{\tau _{X}\})$. Given a causal model $\displaystyle \mathcal{M}$, we define $\displaystyle \mathcal{M}_{V\leftarrow v}( x)$ to be the output of $\displaystyle \mathcal{M}$ with input $\displaystyle x$ after setting $\displaystyle V = v$. For $\displaystyle \mathcal{M}$, where $\{s_{k}\}$ is a source input setting and $\{V_k\}$ is a set of intermediate variables, the interchange intervention (\textit{II}) is defined as: 
\begin{align*}
\textit{II}(\mathcal{M} ,\{s_{k}\} ,\{V_{k}\})( b) & =\mathcal{M}_{V_{k}\leftarrow \text{GetVals}_{V_{k}}(\mathcal{M}( s_{k}))}( b) ,
\end{align*}
where $\displaystyle \text{GetVals}_{V_k}(\mathcal{M}( s_k))$ is the value of variables $\displaystyle V$ in model $\displaystyle \mathcal{M}( s)$. Similarly, one can define \textit{II} with multiple sources as in \cite{geiger2023causal,geiger-etal-2023-DAS}. Ideally, we want the alignment $\displaystyle \Pi $ to satisfy 
\begin{align*}
\tau ( \textit{II}(\mathcal{L} ,s,\pi_X)( b)) & =\textit{II}(\mathcal{H} ,\tau ( s) ,X)( \tau ( b)) .
\end{align*}


\textbf{Distributed Alignment Search (DAS).} Standard-basis interventions implicitly assume that a high-level variable is localized in a particular set of neurons, but neural representations are often distributed across overlapping directions \citep{smolensky1986neural,olah2020zoom,Scherlis2022,geiger-etal-2023-DAS}. Distributed Alignment Search (DAS) addresses this by replacing ``localist'' alignment search with gradient-based search over distributed linear subspaces. Concretely, for a high-level variable \(X\) and a candidate low-level representation \(\mathbf h_{\pi_X}\in\mathbb R^n\), DAS learns an orthogonal matrix \(R_\theta\) and aligns \(X\) with a \(k\)-dimensional subspace of the rotated representation \(R_\theta(\mathbf h_{\pi_X})\), rather than with coordinates in the standard basis \citep{geiger-etal-2023-DAS}. A distributed interchange intervention rotates the base and source representations, swaps the aligned subspace from source into base, and rotates back before continuing the forward pass. The low-level and high-level models are kept fixed; only the alignment parameters are trained so that the intervened low-level output matches the high-level intervention on \(X\). As a variant of DAS, boundless DAS keeps the same objective of searching for better alignments, but replaces the hand-specified subspace size with learned soft boundaries, making the method more scalable \citep{wu-etal-2023-Boundless-DAS}.

\textbf{Sparse Autoencoders (SAEs).} To decompose the dense and polysemantic representations of a language model into interpretable components \citep{elhage2022toy}, we leverage Sparse Autoencoders (SAEs) \citep{bricken2023towards, cunningham2023sparse}. An SAE provides a method for mapping an activation vector $x \in \mathbb{R}^d$ to a high-dimensional, sparse feature space $f(x) \in \mathbb{R}^m$ (where $m \gg d$) via an encoder $f(x) = \text{ReLU}(W_{enc}x + b_{enc})$, such that the original activation can be reconstructed as $\hat{x} = W_{dec}f(x) + b_{dec}$. By training with an $L_1$ penalty to enforce sparsity, SAEs identify features that often correspond to discrete semantic or syntactic concepts \citep{huang2024ravel}. In our diagnostic pipeline, we use SAEs to extract model-internal features for our classifiers.

\section{Finding interchange-consistent Subsets}
\label{sec:method}

Given a causal abstraction candidate \((\mathcal L,\mathcal H,\Pi)\) and an input space \(\mathcal I\), we aim to identify regions of the input space on which the abstraction is fully faithful. Rather than summarizing abstraction quality with a single global IIA score, we seek a more structured view: which inputs are well interpreted by the current hypothesis, which are not, and how this distinction can be used to improve the hypothesis itself. This section proceeds in two parts. Section~\ref{subsec:defining-perfect} formalizes interchange consistency at the level of input pairs and input subsets, yielding a graph-theoretic view of abstraction success and failure \citep{geiger-etal-2020-neural,pislar2025combining}. Section~\ref{subsec:diagnosis-pipeline} then introduces a practical four-step diagnosis pipeline based on this structure to partition the input space and generalize the result beyond the analyzed sample.

\subsection{interchange-consistent Input Subset and the Interchangeability Graph}
\label{subsec:defining-perfect}

Causal abstractions are typically evaluated using interchange intervention accuracy (IIA). Given a causal abstraction \((\mathcal L,\mathcal H,\Pi)\) and a finite set of input pairs \(\mathcal P\), IIA is the proportion of pairs \((i_1,i_2)\in\mathcal P\) for which the low-level intervention matches the corresponding high-level counterfactual for every variable in \(\mathcal H\) (i.e., \(X\in\mathsf{Var}(\mathcal H)\)):
\begin{align*}
\tau\!\left( \text{II}(\mathcal L,i_1,\pi_X)(i_2)\right)
&=
\text{II}\!\left(\mathcal H,\tau(i_1),X\right)\!\left(\tau(i_2)\right).
\end{align*}
While IIA is a useful global summary, it does not reveal how abstraction failures are distributed over the input space. To move beyond a single scalar score, we define a pairwise notion of exact success under interchange interventions and then lift it to subsets of inputs.
\begin{definition}[Interchange-Consistent Pairs]
\label{def:perfect-pairs}
Given a causal abstraction \((\mathcal L,\mathcal H,\Pi)\) and an input set \(\mathcal I\), two inputs \(i_1,i_2\in\mathcal I\) are \emph{interchange-consistent} under \((\mathcal L,\mathcal H,\Pi)\), denoted \(\mathbf p_{(\mathcal H,\mathcal L,\Pi)}\langle i_1,i_2\rangle\), iff for all \(X\in\mathsf{Var}(\mathcal H)\),
\begin{equation*}\label{eq:perfect}
\begin{split}
\tau\!\left( \text{II}(\mathcal L,i_1,\pi_X)(i_2)\right)
&=
\text{II}\left(\mathcal H,\tau(i_1),X\right)\!\left(\tau(i_2)\right),\\
\tau\!\left( \text{II}(\mathcal L,i_2,\pi_X)(i_1)\right)
&=
\text{II}\left(\mathcal H,\tau(i_2),X\right)\left(\tau(i_1)\right).
\end{split}
\end{equation*}
\end{definition}

Definition~\ref{def:perfect-pairs} requires exact counterfactual agreement in both directions: patching from \(i_1\) into \(i_2\), and from \(i_2\) into \(i_1\). We then lift this pairwise notion to subsets.

\begin{definition}[Interchange-Consistent Input Subset]
\label{def:perfect-subspace-method}
An input subset \(I \subseteq \mathcal I\) is \emph{interchange-consistent} by \((\mathcal L,\mathcal H,\Pi)\) iff for all \(i_1,i_2\in I\), \(\mathbf p_{(\mathcal H,\mathcal L,\Pi)}\langle i_1,i_2\rangle\) holds.
\end{definition}

Since we aim to find (quasi-)interchange-consistent input subsets\footnote{In our experiments, we usually search for quasi-interchange-consistent or \(\gamma\)-interchange-consistent input subsets rather than strict interchange-consistent subsets. A subset \(S\subseteq\mathcal I\) is \(\gamma\)-interchange-consistent if at least a \(\gamma\) proportion of input pairs in \(S\) are interchange-consistent. This relaxation reduces computational cost and avoids returning only trivially small subsets when a few failed interventions break an otherwise coherent high-faithfulness region.} we refer to the resulting subsets as \emph{target buckets}, we also refer to such subsets as \emph{target buckets}. This setwise notion naturally admits a graph-theoretic reformulation.

\begin{definition}[Interchangeability Graph]
\label{def:interchangeability-graph}
Given a causal abstraction \((\mathcal L,\mathcal H,\Pi)\) and an input set \(\mathcal I\), the \emph{interchangeability graph} is an undirected graph \(G=(V,E)\) with \(V=\mathcal I\). An edge connects two distinct vertices \(i_1,i_2\in V\) iff \(\mathbf p_{(\mathcal H,\mathcal L,\Pi)}\langle i_1,i_2\rangle\) holds.
\end{definition}

The interchangeability graph provides a map of where a candidate abstraction succeeds. A subset \(I\subseteq \mathcal I\) is interchange-consistent iff every pair of vertices in \(I\) is connected by an edge; in graph-theoretic terms, such a subset is a clique.

\begin{definition}[Clique and Maximum Clique]
\label{def:clique}
A set of vertices \(C\subseteq V\) in an undirected graph \(G=(V,E)\) is a \emph{clique} if every pair of vertices in \(C\) is connected by an edge. A \emph{maximum clique} is a clique of the largest possible size in \(G\).
\end{definition}

One immediate consequence is that a subset \(I\subseteq \mathcal I\) is interchange-consistent by \((\mathcal L,\mathcal H,\Pi)\) iff \(I\) forms a clique in the Interchangeability Graph. This reformulates the search for an exactly faithful region of the input space as a graph problem. 


\subsection{The Diagnosis Pipeline}
\label{subsec:diagnosis-pipeline}

\begin{figure}[t]
    \centering
    \includegraphics[width=.85\linewidth]{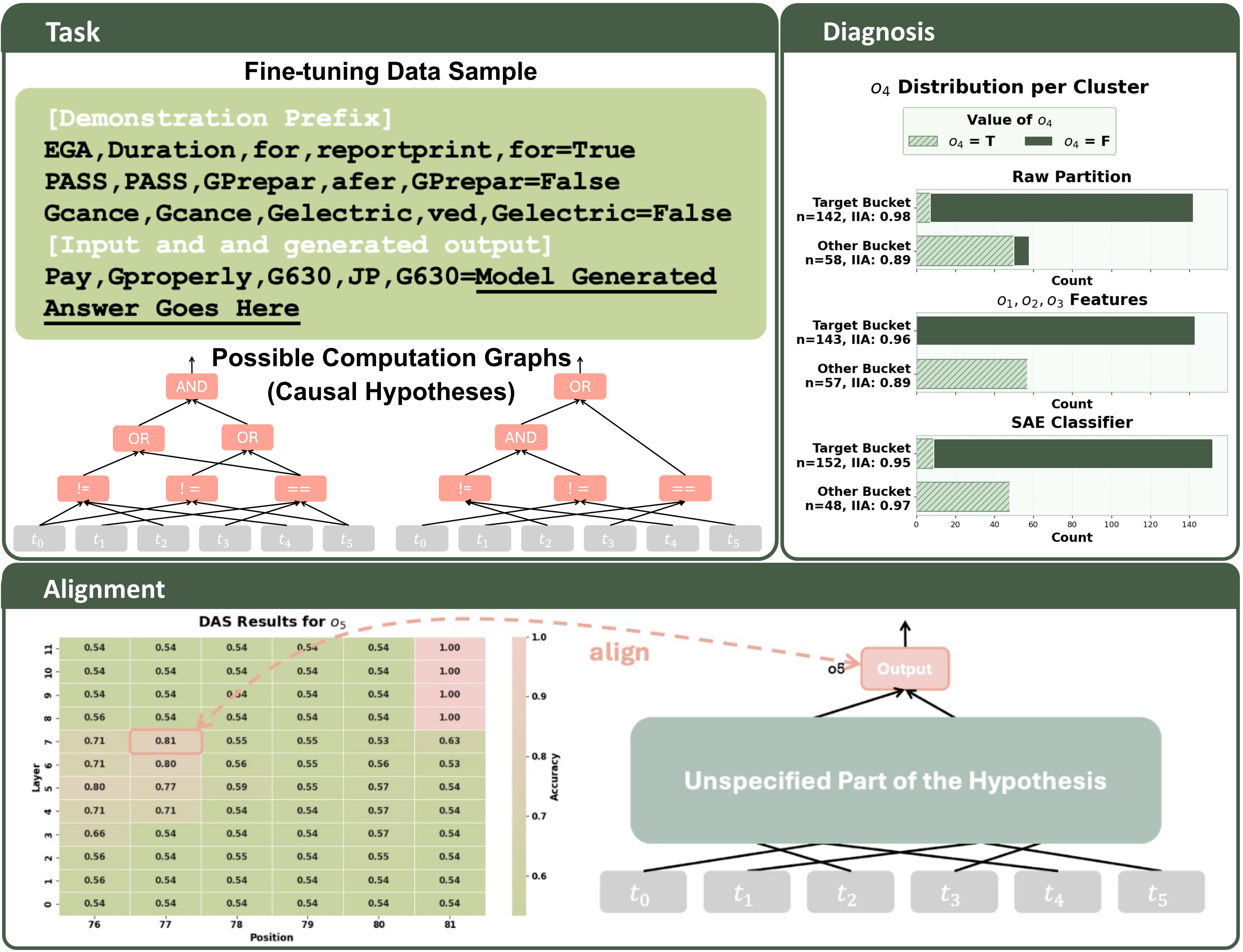}
    \caption{\textbf{Running example -- a toy logic task.} We fine-tune a 12-layer GPT-2-small model to predict the truth value of \(o_5=((t_2\neq t_4)\wedge(t_0\neq t_5))\vee(t_1=t_3)\). For exposition, we denote the primitive Boolean variables (the (non)equalities) by \(o_1,o_2,o_3\), so that the target computation can be written as \(o_5=(o_1\wedge o_2)\vee o_3\). The model is trained only on input-output pairs. This means the model can use different computations to get the same correct result (e.g., $y = (o_1 \wedge o_2) \vee o_3$ vs. $y = (o_1 \vee o_3) \wedge (o_2 \vee o_3)$). Therefore, the task serves as a clean testbed for asking which computation the model is actually using to solve the task.}
    \label{fig:logic-task-first-pass}
\end{figure}

We present a four-step diagnosis pipeline for turning a non-perfect causal abstraction into a more informative object of analysis. The central idea is to move from a single global IIA score to a partition of the input space: rather than asking only whether a hypothesis works on average, we ask where it works, where it fails, and what distinguishes the two regions. The pipeline is task-agnostic, but to make each step concrete, we illustrate it with a running toy logic example (Fig. \ref{fig:logic-task-first-pass}). We use this example here only to show how a \emph{single} diagnosis pass works for one target variable; in Section~\ref{sec:logic-task-recursive}, we return to the same task and show how recursively reapplying the pipeline can recover a fuller high-level hypothesis. The complete diagnosis algorithm is shown in \cref{alg:diagnosis} in the appendix. 


\textbf{Step 1: Specify the task and the input space.}
The first step of the pipeline is to define a task and an input space \(\mathcal I\) on which the low-level model \(\mathcal L\) performs reliably. This isolates failures of \emph{interpretation} from ordinary task failures. In practice, we therefore restrict attention to the subset of correct predictions: $\mathcal I_{\mathrm{correct}}
\;=\;
\{\, i \in \mathcal I \;:\; \mathcal L(i)\ \text{is task-correct}\,\}.$

\textit{Example:} In the toy logic task, we fine-tuned the model on a generated dataset of size \(20,000\), where \(o_1,o_2,o_3\) are each true with probability \(0.5\). The model achieves \(99.7\%\) test accuracy. We filtered out all the failure cases, and the input space consists of all instances where the model correctly predicts the truth value of \(o_5\). 

\textbf{Step 2: Find a candidate alignment.}
Next, given a high-level variable \(X\), we identify a candidate alignment \(\pi_X\) between \(X\) and an intervenable low-level feature. This can be done using any standard alignment-search method, such as DAS or full-vector patching. At this stage, global IIA provides only a preliminary signal: if it is near-perfect, the hypothesis may already be adequate; if it is at chance, the candidate alignment is likely uninformative; but if it is intermediate, the result will need some further investigation.

\textit{Example:} We deliberately begin from an underspecified high-level hypothesis containing only the final output variable \(o_5\). DAS finds a near-perfect alignment at the final output position, but also an earlier candidate alignment at position \(77\), layer \(7\), with IIA approximately \(0.78\). This is precisely the kind of intermediate result that motivates diagnosis: it is clearly above chance, yet far from sufficient to claim a globally faithful encoding of \(o_5\).

\textbf{Step 3: Bucket the input space by interchangeability.}
We then ask whether a non-perfect alignment is faithful on some \emph{subset} of the input space. To answer this, we construct the interchangeability graph over \(\mathcal I_{\mathrm{correct}}\), where two inputs are connected when they are interchange-consistent under the candidate abstraction. Rather than insisting on an exact maximum clique, in practice, we may relax the fully connected requirement and search for $\gamma$-quasi-cliques, which are subgraphs that have density larger than $\gamma$. We choose $\gamma = 0.98$ in all experiments. We treat the dense quasi-clique as the ``target bucket'' and treat the remainder as ``other bucket''. See Appendix~\ref{appendix:method} for details of the bucketing algorithm.

\textit{Example:} Applying this step to the candidate \(o_5\) alignment reveals that the alignment is not merely noisy. The well-interpreted bucket is dominated by inputs for which \(
o_1 \wedge o_2 = \mathrm{False}\), so that the target computation simplifies to \(o_5 = o_3\). On this subspace, a feature that tracks only the \(o_3\) branch can still appear faithful to the full output variable. By contrast, the under-interpreted bucket is dominated by inputs for which \(o_1 \wedge o_2=\mathrm{True}\), where that simplification no longer holds. The diagnosis therefore localizes a specific failure mode of the coarse output-only hypothesis.

\textbf{Step 4: Generalize and characterize the partition.}
Finally, we train a classifier \(g:\mathcal I \to \{1,\dots,K\}\) to predict bucket membership for unseen inputs. The input to the classifier is a feature representation of each input, which can either be hand-labeled features describing the input, or model-internal features such as SAE, activations, or attribution-based features at the aligned site. The output is a bucket label indicating which bucket the input belongs to. This step serves two purposes: it tests whether the discovered buckets reflect a genuine structural boundary rather than an artifact of a finite intervention graph, and it helps characterize that boundary in terms of the features that best predict membership.

\textit{Example:} In the toy logic task, after bucketing the input space, we train two classifiers using bucket membership as the label. One takes the truth values of \(o_1,o_2,o_3\) as hand-labeled input features, and the other takes SAE features at the aligned position as model-internal input features. Both classifiers can predict the membership of unseen inputs, and their predictions are largely consistent. This indicates that the discovered partition reflects a stable structural distinction, and in particular that membership is indeed governed by whether the conjunction branch \(o_4=o_1\wedge o_2\) is active.



Taken together, these four steps turn a single imperfect IIA score into a structured diagnosis. By deliberately aligning the output variable to a location with an imperfect but above baseline IIA, we identify buckets of inputs with high in-bucket and low cross-bucket IIA. This shows that this alignment is locally faithful within each bucket but collapses distinctions across buckets, indicating that the output variable can be further broken down into two intermediate variables. Figure~\ref{fig:logic-task-first-pass} illustrates this first diagnosis pass on the toy logic task.

\section{Experiments}

We evaluate the diagnosis recipe in three settings of increasing complexity: a synthetic logic task, an entity-binding task adapted from prior work \citep{gur2025mixing}, and an entangled factual recall setting based on the RAVEL benchmark \citep{huang2024ravel}. In each case, we begin with an existing causal abstraction obtained using a standard alignment method, observe that its global IIA is informative but non-perfect, and then apply our partitioning procedure to identify buckets with high within-bucket and low cross-bucket IIA. We finally test whether these partitions generalize beyond the analyzed sample and whether they suggest useful refinements to the original abstraction.

\subsection{Recursive Hypothesis Discovery in the Toy Logic Task}
\label{sec:logic-task-recursive}

Section~\ref{subsec:diagnosis-pipeline} established the first diagnosis pass for the toy logic task. Aligning \(o_5\) to position \(77\), layer \(7\) partitions the input space into two buckets, both of which have high within-bucket IIA but low cross-bucket IIA. The partition has a clear semantic interpretation: in the target bucket, \(o_4=o_1\wedge o_2\) is always \texttt{False}, whereas in the other bucket, \(o_4\) is always \texttt{True}. This shows that the candidate \(o_5\) alignment does not represent the full output uniformly. Instead, it separates two regimes of computation, suggesting that \(o_5\) should be refined into two variables, \(o_4\) and \(o_3\).

\begin{figure}[t]
    \centering
    \includegraphics[width=1\linewidth]{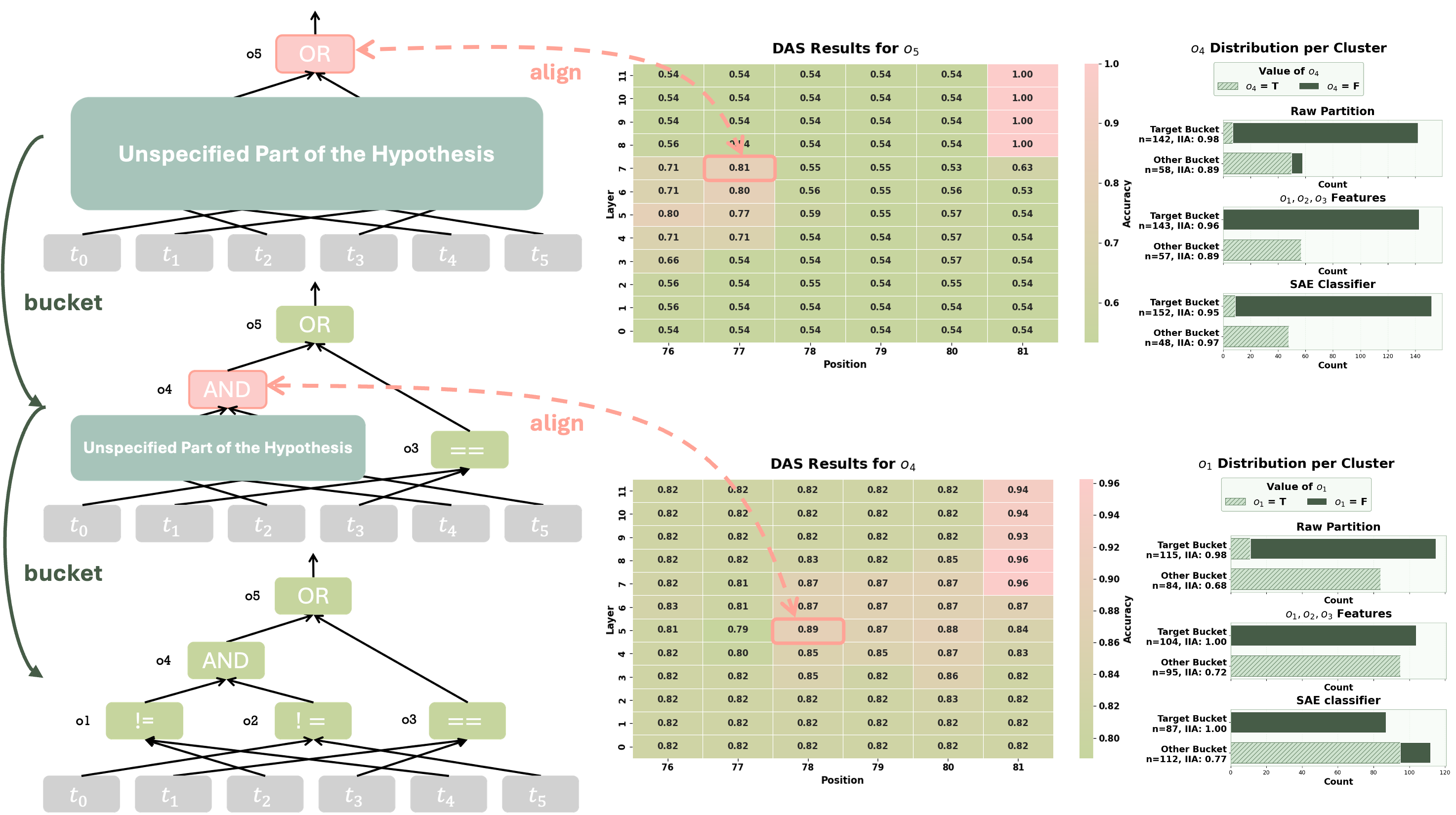}
    \caption{\textbf{Recursive hypothesis discovery in the toy logic task.} \textbf{Top:} Diagnosing the candidate \(o_5\) alignment at position \(77\), layer \(7\) partitions the input space into two high-IIA buckets separated by the latent variable \(o_4=o_1\wedge o_2\). \textbf{Middle:} After promoting \(o_4\) to an explicit variable, DAS identifies both a near-perfect signal at position \(81\), layer \(7\) and a non-trivial earlier signal at position \(78\), layer \(5\). \textbf{Bottom:} Diagnosing the earlier \(o_4\) signal reveals \(o_1\) as the next missing component, recovering the complete hierarchy.}
    \label{fig:logic_recur}
\end{figure}

We then apply the same diagnosis procedure to \(o_4\). DAS identifies a near-perfect signal for \(o_4\) at position \(81\), layer \(7\), indicating that the completed \(o_4\) computation can be localized there, and also a non-trivial above-baseline signal at position \(78\), layer \(5\). We align \(o_4\) to this earlier site and repeat the bucketing step. The resulting partition again yields two buckets with high within-bucket IIA, and in the target bucket \(o_1\) is always \texttt{False}. This shows that the earlier \(o_4\) signal can itself be decomposed into the primitive components \(o_1\) and \(o_2\). Classifiers trained on both hand-labeled features and SAE features generalize these bucket assignments well, and their predictions are largely consistent, indicating that the discovered boundaries reflect stable structure rather than artifacts of a finite intervention graph.

Taken together, these two passes recover the hierarchy \(o_1,o_2,o_3 \rightarrow o_4 \rightarrow o_5\). Starting from the output variable alone, recursive diagnosis spells out the full high-level causal hypothesis from top to bottom. In particular, it shows that the model solves the task via the intended factorization \(y=(o_1\wedge o_2)\vee o_3\), rather than the extensionally equivalent alternative \(y=(o_1\vee o_3)\wedge(o_2\vee o_3)\). Consistent with this picture, alignment search over all variables localizes \(o_3\) at position \(77\), layer \(7\), and \(o_1\) at position \(78\), layer \(5\) (Appendix~\ref{appendix:logic}).

\subsection{Entity Binding Task}

\begin{figure}[!ht]
    \centering
    \includegraphics[width=\linewidth]{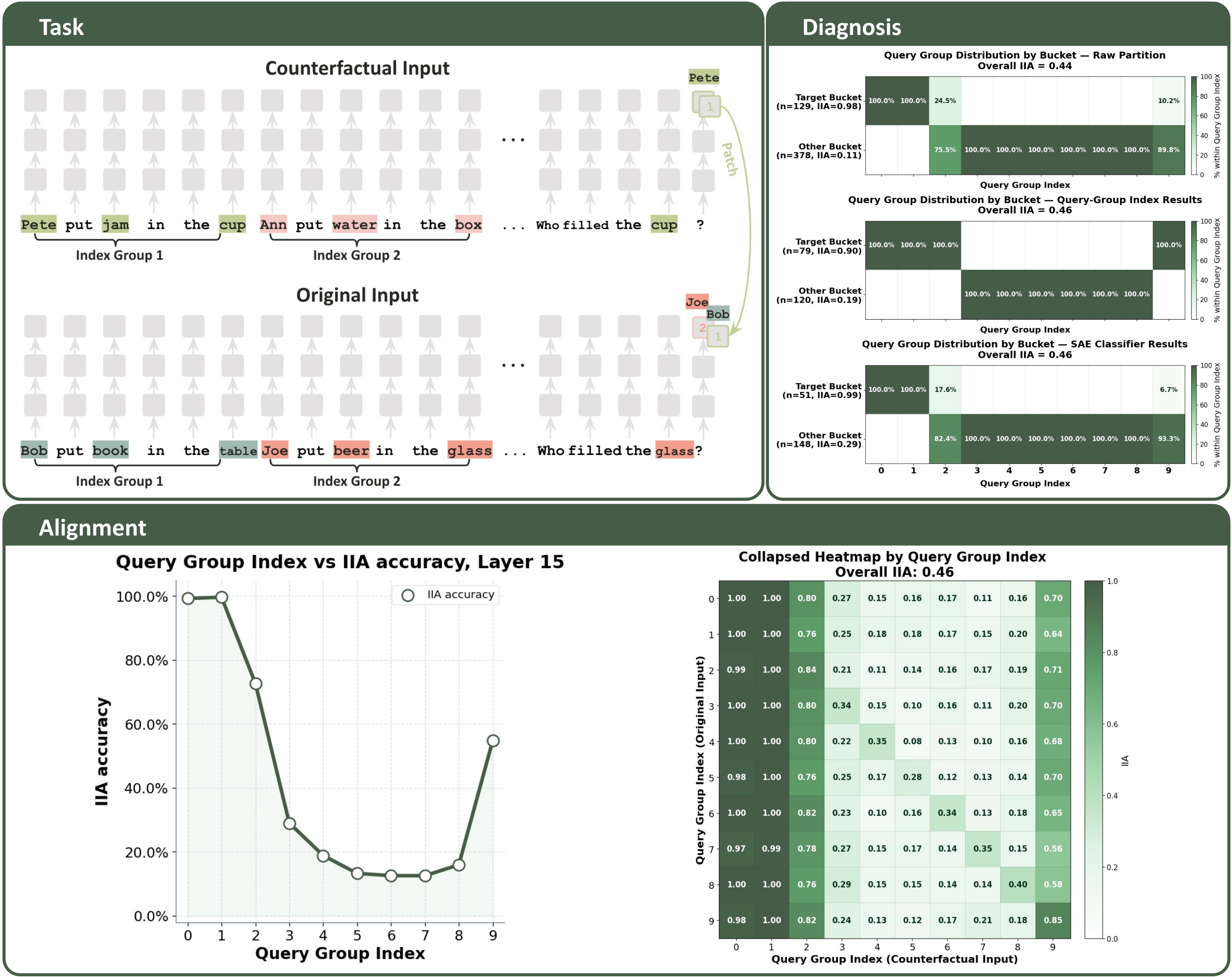}
    \caption{\textbf{Diagnosis of the Entity Binding Task in Gemma-2-2B-Instruct.} \textbf{Top Left:} The entity-binding task evaluates in-context retrieval from sequences of templated groups (e.g., ``John fills a cup with beer...''). A query (e.g., ``Who filled a cup?'') then requests an entity from a specific group. We test a \emph{positional hypothesis}: retrieval is mediated by a high-level variable $q_{group}$ representing the queried group’s context position. The model must identify this position and dereference it to retrieve the target entity. \textbf{Bottom:} Preliminary evaluation via full-vector patching at layer 15 reveals a U-shaped alignment curve as in \citet{gur2025mixing}. \textbf{Top Right:} Our procedure constructs an interchangeability graph to isolate the well-interpreted and under-interpreted regions. This principled diagnosis recovers the failure mode automatically: the target bucket is dominated by edge groups ($q_{group} \in \{0, 1, 2, 9\}$), while the other bucket contains medial groups ($q_{group} \in \{3, 4, 5, 6, 7, 8\}$). }
    \label{fig:Entity Binding}
\end{figure}

\textbf{Task and input space.}
We evaluate our framework using the entity-binding task \citep{geiger2020neural,gur2025mixing}, where a model must retrieve specific entities from a sequence of templatic \emph{entity groups}. We test the \emph{positional hypothesis}—the idea that retrieval is mediated by a high-level variable $q_{\mathrm{group}}$ representing the context position of the queried group—using the \texttt{filling\_liquids} task with \texttt{Gemma-2-2B-Instruct}. Working within a 10-group input space $\mathcal{I}$ (where task accuracy exceeds $98\%$), we leverage this setup (Fig.~\ref{fig:Entity Binding}) to diagnose the model's internal positional mechanism. The input space $\mathcal{I}$ comprises these correctly answered prompts, annotated by the queried group position and the fixed within-group roles.



\textbf{Alignment and Preliminary Evaluation.}
We replicate the positional-hypothesis experiment from \citet{gur2025mixing}, defining a high-level model where a single variable, $q_{\mathrm{group}}$, represents the queried group's position. We localize the corresponding low-level site using vanilla patching and evaluate the hypothesis via full-vector patching at the final query position. Our preliminary results at layer 15 reveal a sharply non-uniform alignment curve, shown in Fig~\ref{fig:Entity Binding} , successfully reproducing the U-shape reported by \cite{gur2025mixing}. The positional mechanism is highly faithful for entity groups at the sequence's edges but degrades significantly for those in the middle. Because the resulting global IIA is neither near-perfect nor near-chance, the aggregate scalar summary fails to characterize the specific input subspaces where the interpretation remains valid. This ambiguity underscores the need for a diagnostic procedure to isolate the well-interpreted regions of the input space.

\textbf{Bucketing the Input Space.}
We apply our partitioning procedure to the positional abstraction $(\mathcal L, \mathcal H_{\mathrm{pos}}, \Pi)$. The overall density of the interchangeability graph is $24\%$, signaling that the abstraction is far from uniformly faithful across the entire input space. Upon partitioning this graph, a clear semantic pattern emerges (Fig~\ref{fig:Entity Binding} Upper Right). The well-interpreted bucket is dominated by inputs where the queried group is located at the start or end of the sequence, while the bad bucket is dominated by inputs in the middle. This diagnosis recovers the failure mode identified manually in \citet{gur2025mixing}. The strength of our method is that this conclusion emerges directly from the structure of the intervention graph itself. Rather than manually inspecting per-index intervention curves, we obtain a principled partition of the input space into well-interpreted and under-interpreted regions. 

\textbf{Generalization through classifier.}
To test whether this partition generalizes beyond the finite graph used for bucketing, we train classifiers to predict bucket membership for unseen in-distribution inputs. We train classifiers on both the queried group index and internal SAE features. The index-based classifier recovers an interpretable rule—assigning $q_{\mathrm{group}} \in \{0,1,2,9\}$ to the target bucket and $\{3, \dots, 8\}$ to the other bucket—attaining $90\%$ IIA with $81\%$ density (Fig~\ref{fig:Entity Binding}, middle right). The SAE-based classifier further improves this to $98\%$ IIA at $97\%$ density (Fig~\ref{fig:Entity Binding}, bottom right). That internal features outperform hand-labeled features suggests the distinction between inputs in different buckets is fundamentally encoded within the model's own representations.

\textbf{Takeaway.}
This experiment demonstrates our framework's ability to transform known mechanistic failures into reusable diagnostic objects. While the initial steps replicate the positional abstraction in \cite{gur2025mixing}, our bucketing procedure reveals that the hypothesis is not merely "imperfect" on average, but highly faithful on a structured subset while systematically unfaithful elsewhere. By generalizing this boundary to unseen inputs via classifiers, we find that internal SAE features outperform hand-labeled features. This recipe moves beyond reproducing known failure modes to effectively localize, characterize, and operationalize the boundaries of model interpretations.

\subsection{Entangled Factual Recall}

\begin{figure}[t]
    \centering
    \includegraphics[width=1\linewidth]{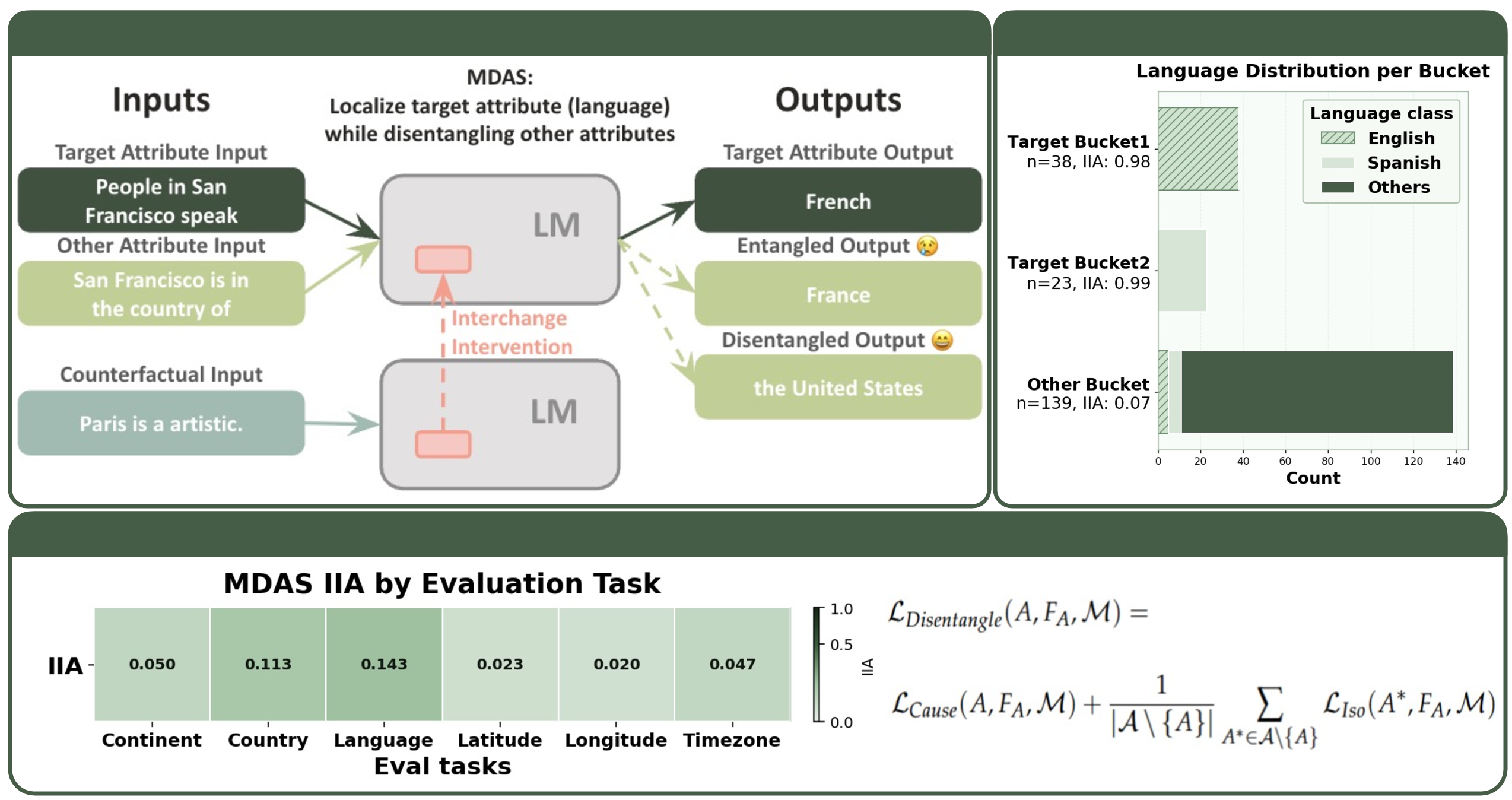}
    \caption{\textbf{Entangled Factual Recall Task and Diagnosis.} \textbf{Top Left:} This task utilizes the RAVEL benchmark to evaluate whether a specific entity attribute (e.g., \textsc{Language}) can be isolated from other co-encoded attributes like \textsc{Country}, \textsc{Continent}, and \textsc{Timezone}. It is particularly challenging because these features are often entangled within a single internal model state, making surgical intervention difficult. \textbf{Bottom:} MDAS Results. We train the alignment using all samples from all attributes to isolate the \textsc{Language} attribute and evaluate it on all attributes.  \textbf{Top Right:} After partition, the "Target Buckets" correspond to regions where the representation is highly faithful for specific languages—English and Spanish—while the "Other Bucket" captures the remaining modes.}
    \label{fig:factual}
\end{figure}

\textbf{Task and input space.}
We evaluate our framework on \emph{entangled factual recall} using the RAVEL benchmark \citep{huang2024ravel}. This task tests the ability of interpretability methods to isolate a specific entity attribute from others co-encoded in the same representation (\cref{fig:factual}). This task is particularly challenging because attributes are often entangled within a single internal state, making it difficult to intervene on one without affecting others. Using a pretrained \texttt{Llama-3.1-8B} model, we focus on isolating the \textsc{Language} attribute from others. 

\textbf{Alignment and Preliminary Evaluation}
A successful alignment in the \textsc{Ravel} benchmark must satisfy two key properties: {causal effectiveness} and {isolation}.For a target attribute $X \in \mathcal{A}$ and its aligned representation $\Pi_X$ mapped via $\tau_X$, a high \textbf{Cause} score signifies that an interchange intervention $II(\mathcal{M}, s, \Pi_X)(b)$ successfully propagates the attribute value $\tau_X(s)$ from the source $s$ into the model's output for the base input $b$. Conversely, a high \textbf{Iso} (isolation) score ensures the intervention is surgically precise, remaining invariant to the states of extraneous features $Y \in \mathcal{A} \setminus \{X\}$. The formal loss objectives are detailed in (\ref{eq:loss_cause}) and (\ref{eq:loss_iso}) in the appendix. While Multi-task Distributed Alignment Search (MDAS) is architected to jointly optimize these criteria, the resulting alignment yields a held-out Interchange Intervention Accuracy (IIA) of only 14.3\%. This discrepancy identifies a central theoretical puzzle: despite MDAS being explicitly designed for attribute disentanglement , the learned subspace remains a fragile abstraction. This failure points to a structural mismatch that aggregate metrics cannot illuminate, necessitating our partitioning method to diagnose why the alignment fails to generalize beyond the training distribution

\textbf{Bucketing the input space.}
We next apply our bucketing procedure to the learned MDAS language subspace. The resulting intervention  graph is sparse overall, with edge density only about \(10\%\). Once we partition this graph, however, a clear structure emerges (Fig~\ref{fig:factual} right). The dense buckets correspond almost exactly to single-language regions: one 
bucket is dominated by English-speaking cities and Spanish-speaking cities, each with high IIA (around \(98 \%\)). The remaining cities fall into a diffuse ``other languages'' bucket with very low IIA. This shows that the learned MDAS subspace does not faithfully represent the \textsc{Language} attribute in general. Instead, it collapses the input space into a small 
number of coarse language-specific clusters, preserving interchangeability mainly within the same language while failing on cross-language interventions. 

\textbf{Generalization through classifier.}
To test generalization, we train an SAE-based classifier to predict bucket membership for unseen inputs. The classifier reproduces the partition with $90\%$ accuracy, confirming that the language-specific structure is indeed encoded in the model's internal representations. Inspecting the highest-weight SAE features reveals only weak signals that are loosely consistent with the corresponding languages. These signals are far less direct than the buckets themselves: while reading SAE feature descriptions provides only weak, post hoc evidence, examining the grouped inputs in each bucket makes the underlying structure immediately clear.



\textbf{Takeaway.}
This experiment reveals a limitation of MDAS-style disentanglement in the entangled factual recall setting. Although MDAS is designed to isolate the target concept from correlated non-target attributes, our analysis shows that it can do so only by \emph{collapsing the target representation} itself. The resulting subspace is partially isolated but only weakly faithful: it preserves coarse within-language structure while losing the finer-grained information needed for robust cross-language interchangeability. In this case, bucketing makes the failure mode explicit by showing that the learned representation separates a few major language clusters rather than capturing the \textsc{Language} attribute as a whole.

\vspace{-0.5 em}
\section{Conclusion and Future Work}

\vspace{-0.5 em}
We introduced a method for diagnosing and improving causal abstractions by bucketing the input space according to interchange-intervention behavior. Rather than treating IIA as a single global summary, our method identifies subsets of inputs on which a proposed abstraction is nearly perfect. This turns causal abstraction into a more diagnostic and constructive framework
. Across fine-tuned and pretrained models of different sizes, and across alignment methods including DAS, MDAS, and full-vector patching, our experiments show that bucketing is a broadly applicable diagnostic tool for causal abstraction; in the toy logic task, it also supports iterative discovery of the high-level hypothesis itself.

Our approach has two main limitations. First, it depends on a prior task construal: it operates within a specified task and candidate abstraction, rather than performing unconstrained automatic interpretation. In this sense, it is better understood as a structured program search than as a general solution to mechanistic interpretability. Second, our experiments focus on relatively simple, mostly single-variable hypotheses. In more complex settings, interacting variables and accumulated approximation error may make it difficult to discover large buckets with high IIA. 
These limitations suggest two natural directions for future work. One is to scale the method to richer tasks, larger input spaces, and higher-dimensional or genuinely multi-variable abstractions. Another is to better understand how bucketing can support more automated hypothesis discovery and refinement in non-trivial settings---for example, by turning the structure of identified buckets into a systematic search procedure over candidate variables, decompositions, and compositions of partial hypotheses.

\section*{Acknowledgements}

We thank Zhengxuan Wu for extensive guidance and support during the exploration phase of this project. We also thank Zhengxuan Wu, Nathan Roll, and Amir Zur for helpful comments and suggestions on earlier drafts of this paper.

\bibliographystyle{plainnat}
\bibliography{references}

@article{geiger2021causal,
  title={Causal abstractions of neural networks},
  author={Geiger, Atticus and Lu, Hanson and Icard, Thomas and Potts, Christopher},
  journal={Advances in Neural Information Processing Systems},
  volume={34},
  pages={9574--9586},
  year={2021}
}

@article{Scherlis2022,
  author       = {Adam Scherlis and
                  Kshitij Sachan and
                  Adam S. Jermyn and
                  Joe Benton and
                  Buck Shlegeris},
  title        = {Polysemanticity and Capacity in Neural Networks},
  journal      = {CoRR},
  volume       = {abs/2210.01892},
  year         = {2022},
  url          = {https://doi.org/10.48550/arXiv.2210.01892},
  doi          = {10.48550/ARXIV.2210.01892},
  eprinttype    = {arXiv},
  eprint       = {2210.01892},
  bibsource    = {dblp computer science bibliography, https://dblp.org}
}

@article{olah2020zoom,
  author = {Olah, Chris and Cammarata, Nick and Schubert, Ludwig and Goh, Gabriel and Petrov, Michael and Carter, Shan},
  title = {Zoom In: An Introduction to Circuits},
  journal = {Distill},
  year = {2020},
  note = {https://distill.pub/2020/circuits/zoom-in},
  doi = {10.23915/distill.00024.001}
}

@article{wu2024pyvene,
  title={pyvene: A library for understanding and improving PyTorch models via interventions},
  author={Wu, Zhengxuan and Geiger, Atticus and Arora, Aryaman and Huang, Jing and Wang, Zheng and Goodman, Noah D and Manning, Christopher D and Potts, Christopher},
  journal={arXiv preprint arXiv:2403.07809},
  year={2024}
}

@inproceedings{sundararajan2017axiomatic,
  title={Axiomatic attribution for deep networks},
  author={Sundararajan, Mukund and Taly, Ankur and Yan, Qiqi},
  booktitle={International conference on machine learning},
  pages={3319--3328},
  year={2017},
  organization={PMLR}
}

@article{geiger2023causal,
  title={Causal abstraction: A theoretical foundation for mechanistic interpretability},
  author={Geiger, Atticus and Ibeling, Duligur and Zur, Amir and Chaudhary, Maheep and Chauhan, Sonakshi and Huang, Jing and Arora, Aryaman and Wu, Zhengxuan and Goodman, Noah and Potts, Christopher and others},
  journal={arXiv preprint arXiv:2301.04709},
  year={2023}
}

@inproceedings{geiger-etal-2020-neural,
	address = {Online},
	author = {Geiger, Atticus and Richardson, Kyle and Potts, Christopher},
	booktitle = {Proceedings of the Third BlackboxNLP Workshop on Analyzing and Interpreting Neural Networks for NLP},
	doi = {10.18653/v1/2020.blackboxnlp-1.16},
	month = nov,
	pages = {163--173},
	publisher = {Association for Computational Linguistics},
	title = {Neural Natural Language Inference Models Partially Embed Theories of Lexical Entailment and Negation},
	url = {https://www.aclweb.org/anthology/2020.blackboxnlp-1.16},
	year = {2020}}

@article{meng2022locating,
  title={Locating and Editing Factual Associations in {GPT}},
  author={Kevin Meng and David Bau and Alex Andonian and Yonatan Belinkov},
  journal={Advances in Neural Information Processing Systems},
  volume={36},
  year={2022}
}

@article{geiger-etal-2023-DAS,
    title={Finding Alignments Between Interpretable Causal Variables and Distributed Neural Representations}, 
    author={Geiger, Atticus and Wu, Zhengxuan and Potts, Christopher and Icard, Thomas  and Goodman, Noah},
    year={2023},
    booktitle={arXiv}
}

@article{wu-etal-2023-Boundless-DAS,
    title={Interpretability at Scale: Identifying Causal Mechanisms in Alpaca}, 
    author={Wu, Zhengxuan and Geiger, Atticus and Icard, Thomas and Potts, Christopher and Goodman, Noah},
    year={2023},
    booktitle={NeurIPS}
}

@inproceedings{huanginternal,
  title={Internal Causal Mechanisms Robustly Predict Language Model Out-of-Distribution Behaviors},
year={2025},
  author={Huang, Jing and Tao, Junyi and Icard, Thomas and Yang, Diyi and Potts, Christopher},
  booktitle={Forty-second International Conference on Machine Learning}
}

@article{gur2025mixing,
  title={Mixing Mechanisms: How Language Models Retrieve Bound Entities In-Context},
  author={Gur-Arieh, Yoav and Geva, Mor and Geiger, Atticus},
  journal={arXiv preprint arXiv:2510.06182},
  year={2025}
}

@article{geiger2025causal,
  title={Causal abstraction: A theoretical foundation for mechanistic interpretability},
  author={Geiger, Atticus and Ibeling, Duligur and Zur, Amir and Chaudhary, Maheep and Chauhan, Sonakshi and Huang, Jing and Arora, Aryaman and Wu, Zhengxuan and Goodman, Noah and Potts, Christopher and others},
  journal={Journal of Machine Learning Research},
  volume={26},
  number={83},
  pages={1--64},
  year={2025}
}

@inproceedings{boguraev-etal-2025-causal,
    title = {Causal Interventions Reveal Shared Structure Across {E}nglish Filler{--}Gap Constructions},
    author = {Boguraev, Sasha and Potts, Christopher and Mahowald, Kyle},
     editor = {Christodoulopoulos, Christos  and Chakraborty, Tanmoy  and Rose, Carolyn  and Peng, Violet},
    booktitle = {Findings of the Association for Computational Linguistics: EMNLP 2025},
    year = {2025},
    address = {Suzhou, China},
    publisher = {Association for Computational Linguistics},
    url = {https://aclanthology.org/2025.emnlp-main.1271/},
    pages = {25032--25053}}

@book{pearl2009causality,
  title={Causality},
  author={Pearl, Judea},
  year={2009},
  publisher={Cambridge university press}
}

@inproceedings{arora-etal-2024-causalgym,
    title = {{C}ausal{G}ym: Benchmarking Causal Interpretability Methods on Linguistic Tasks},
    author = {Arora, Aryaman and Jurafsky, Dan and Potts, Christopher},
    editor = {Ku, Lun-Wei  and Martins, Andre and Srikumar, Vivek},
    booktitle = {Proceedings of the 62nd Annual Meeting of the Association for Computational Linguistics (Volume 1: Long Papers)},
    year = {2024},
    address = {Bangkok, Thailand},
    publisher = {Association for Computational Linguistics},
    url = {https://aclanthology.org/2024.acl-long.785},
    pages = {14638--14663}}

@inproceedings{huang2024ravel,
  title={RAVEL: Evaluating interpretability methods on disentangling language model representations},
  author={Huang, Jing and Wu, Zhengxuan and Potts, Christopher and Geva, Mor and Geiger, Atticus},
  booktitle={Proceedings of the 62nd Annual Meeting of the Association for Computational Linguistics (Volume 1: Long Papers)},
  pages={8669--8687},
  year={2024}
}

@article{bricken2023towards,
  title={Towards Monosemanticity: Decomposing Language Models with Sparse Autoencoders},
  author={Bricken, Adly and Templeton, Adni and Batson, Joshua and Chen, Brian and Jerome, Adam and Moore, Scott and Tamkin, Shahar and Jones, Landon and Conerly, Dustin and Cunningham, Hoagy and others},
  journal={Transformer Circuits Thread},
  year={2023}
}

@article{he2024llamascope,
  title={Llama Scope: Extracting Millions of Features from Llama-3.1-8B with Sparse Autoencoders},
  author={He, Zhengfu and Shu, Wentao and Ge, Xuyang and Chen, Lingjie and Wang, Junxuan and Zhou, Yunhua and Liu, Frances and Guo, Qipeng and Huang, Xuanjing and Wu, Zuxuan and others},
  journal={arXiv preprint arXiv:2410.20526},
  year={2024}
}

@article{lieberum2024gemmascope,
  title={Gemma Scope: Open Sparse Autoencoders Everywhere All At Once on Gemma 2},
  author={Lieberum, Tom and Rajamanoharan, Senthooran and Conmy, Arthur and Smith, Lewis and Sonnerat, Nicolas and Varma, Vikrant and Kram{\'a}r, J{\'a}nos and Dragan, Anca and Shah, Rohin and Nanda, Neel},
  journal={arXiv preprint arXiv:2408.05147},
  year={2024}
}

@article{makelov2023subspace,
  title={Is this the subspace you are looking for? an interpretability illusion for subspace activation patching},
  author={Makelov, Aleksandar and Lange, Georg and Nanda, Neel},
  journal={arXiv preprint arXiv:2311.17030},
  year={2023}
}

@article{meng2022mass,
  title={Mass-editing memory in a transformer},
  author={Meng, Kevin and Sharma, Arnab Sen and Andonian, Alex and Belinkov, Yonatan and Bau, David},
  journal={arXiv preprint arXiv:2210.07229},
  year={2022}
}

@article{hernandez2023linearity,
  title={Linearity of relation decoding in transformer language models},
  author={Hernandez, Evan and Sharma, Arnab Sen and Haklay, Tal and Meng, Kevin and Wattenberg, Martin and Andreas, Jacob and Belinkov, Yonatan and Bau, David},
  journal={arXiv preprint arXiv:2308.09124},
  year={2023}
}

@article{cunningham2023sparse,
  title={Sparse autoencoders find highly interpretable features in language models},
  author={Cunningham, Hoagy and Ewart, Aidan and Riggs, Logan and Huben, Robert and Sharkey, Lee},
  journal={arXiv preprint arXiv:2309.08600},
  year={2023}
}

@article{elhage2022toy,
  title={Toy models of superposition},
  author={Elhage, Nelson and Hume, Tristan and Olsson, Catherine and Schiefer, Nicholas and Henighan, Tom and Kravec, Shauna and Hatfield-Dodds, Zac and Lasenby, Robert and Drain, Dawn and Chen, Carol and others},
  journal={arXiv preprint arXiv:2209.10652},
  year={2022}
}

@inproceedings{geva2021transformer,
  title={Transformer feed-forward layers are key-value memories},
  author={Geva, Mor and Schuster, Roei and Berant, Jonathan and Levy, Omer},
  booktitle={Proceedings of the 2021 Conference on Empirical Methods in Natural Language Processing},
  pages={5484--5495},
  year={2021}
}

@inproceedings{geva2023dissecting,
  title={Dissecting recall of factual associations in auto-regressive language models},
  author={Geva, Mor and Bastings, Jasmijn and Filippova, Katja and Globerson, Amir},
  booktitle={Proceedings of the 2023 Conference on Empirical Methods in Natural Language Processing},
  pages={12216--12235},
  year={2023}
}

@article{meloux2025everything,
  title={Everything, everywhere, all at once: is mechanistic interpretability identifiable?},
  author={M{\'e}loux, Maxime and Maniu, Silviu and Portet, Fran{\c{c}}ois and Peyrard, Maxime},
  journal={arXiv preprint arXiv:2502.20914},
  year={2025}
}

@article{wu2024reply,
  title={A reply to makelov et al.(2023)'s" interpretability illusion" arguments},
  author={Wu, Zhengxuan and Geiger, Atticus and Huang, Jing and Arora, Aryaman and Icard, Thomas and Potts, Christopher and Goodman, Noah D},
  journal={arXiv preprint arXiv:2401.12631},
  year={2024}
}

@article{smolensky1986neural,
  title={Neural and conceptual interpretation of PDP models},
  author={Smolensky, Paul},
  journal={Parallel distributed processing: Explorations in the microstructure of cognition},
  volume={2},
  pages={390--431},
  year={1986},
  publisher={MIT Press/Bradford Books Cambridge, MA}
}

@article{chalupka2017causal,
  title={Causal feature learning: an overview},
  author={Chalupka, Krzysztof and Eberhardt, Frederick and Perona, Pietro},
  journal={Behaviormetrika},
  volume={44},
  number={1},
  pages={137--164},
  year={2017},
  publisher={Springer}
}

@article{chalupka2014visual,
  title={Visual causal feature learning},
  author={Chalupka, Krzysztof and Perona, Pietro and Eberhardt, Frederick},
  journal={arXiv preprint arXiv:1412.2309},
  year={2014}
}

@article{pislar2025combining,
  title={Combining causal models for more accurate abstractions of neural networks},
  author={P{\^\i}slar, Theodora-Mara and Magliacane, Sara and Geiger, Atticus},
  journal={arXiv preprint arXiv:2503.11429},
  year={2025}
}

@inproceedings{geiger2020neural,
  title={Neural natural language inference models partially embed theories of lexical entailment and negation},
  author={Geiger, Atticus and Richardson, Kyle and Potts, Christopher},
  booktitle={Proceedings of the third blackboxnlp workshop on analyzing and interpreting neural networks for NLP},
  pages={163--173},
  year={2020}
}

\appendix
\section{Methodology}
\label{appendix:method}
The formal diagnosis pipeline is shown in \cref{alg:diagnosis} and \cref{alg:multi_seed_quasi}. 

\paragraph{Greedy Multi-Seed Quasi-Clique Search}
As the identification of maximum cliques within the Interchangeability Graph is generally NP-hard, we implement a heuristic greedy search to approximate the interchange-consistent (perfectly interpreted) subspaces (\cref{alg:multi_seed_quasi}).
This algorithm serves as the implementation of the \textsc{FindMaximalQuasiClique} function within our diagnosis pipeline. 

To maximize the probability of discovering large dense regions, the algorithm employs a multi-seed strategy starting from the most promising candidates. %
We first compute the degrees of all nodes in $V \subseteq \mathcal{I}_{\mathrm{correct}}$ based on the current subgraph and sort them in descending order. %
The expansion is then attempted independently for each of the top 10 highest-degree nodes as seeds. %
For a given seed, the algorithm iteratively expands the set $C$ by selecting the candidate node $w$ that maximizes the resulting edge density $\rho$, provided that $\rho$ remains above the user-defined threshold $\gamma$. %

The density threshold $\gamma \in (0, 1]$ allows the framework to be robust to minor representational noise; while $\gamma=1.0$ identifies a perfect clique, a slightly lower value (e.g., $\gamma=0.9$) allows the pipeline to isolate regions of high causal faithfulness that might otherwise be fragmented by trivial neural variations. We choose $\gamma = 0.98$ for all experiments in the paper. 
The algorithm tracks the results of each seed and returns the largest quasi-clique $C_{\mathrm{best}}$, which is then extracted as a ``Target Bucket'' before the search space is updated for the next iteration. %

\begin{algorithm}[ht]
\caption{The Diagnosis Pipeline}
\label{alg:diagnosis}
\begin{algorithmic}[1]
\REQUIRE Causal abstraction $(\mathcal{L}, \mathcal{H}, \Pi)$, input space $\mathcal{I}$, density threshold $\gamma$, max buckets $K$
\STATE $\mathcal{I}_{\mathrm{correct}} \leftarrow \{i \in \mathcal{I} : \mathcal{L}(i) \text{ is correct}\}$
\STATE Construct Interchangeability Graph $G = (V, E)$ on $\mathcal{I}_{\mathrm{correct}}$ using Eq.~\ref{eq:perfect}
\FOR{$j = 1$ \TO $K-1$}
    \STATE $C_j \leftarrow \textsc{FindMaximalQuasiClique}(G, \gamma)$ \hfill \# Identify a dense, faithful region
    \IF{$C_j = \emptyset$}
        \STATE \textbf{break}
    \ENDIF
    \STATE $V \leftarrow V \setminus C_j$ \hfill \# Remove identified region from the search space
\ENDFOR
\STATE $C_K \leftarrow V$ \hfill \# Residual bucket of failure modes
\STATE $\mathcal{D}_{\mathrm{train}} \leftarrow \{(i, j) \mid i \in C_j, j \in \{1, \dots, K\}\}$
\STATE $g \leftarrow \textsc{TrainClassifier}(\mathcal{D}_{\mathrm{train}})$ \hfill \# Learn to generalize the partition
\RETURN Buckets $\{C_j\}_{j=1}^K$ and diagnostic classifier $g$
\end{algorithmic}
\end{algorithm}

\begin{algorithm}[ht]
\caption{Multi-Seed Greedy Quasi-Clique Search}
\label{alg:multi_seed_quasi}
\begin{algorithmic}[1]
\REQUIRE Adjacency matrix $A$, available nodes $V \subseteq \mathcal{I}_{\mathrm{correct}}$, density threshold $\gamma$, minimum size $s$
\IF{$|V| < s$} \RETURN $\emptyset$ \ENDIF
\STATE Compute degrees of nodes in $V$ based on subgraph $G[V]$
\STATE $V_{\mathrm{sorted}} \leftarrow \text{nodes in } V \text{ sorted by degree descending}$
\STATE $C_{\mathrm{best}} \leftarrow \emptyset$
\FOR{$v_{\mathrm{seed}}$ \textbf{in} $V_{\mathrm{sorted}}[1 \dots \min(10, |V|)]$}
    \STATE $C \leftarrow \{v_{\mathrm{seed}}\}$; $S \leftarrow V \setminus \{v_{\mathrm{seed}}\}$
    \STATE $\text{improved} \leftarrow \text{True}$
    \WHILE{$\text{improved}$ \AND $S \neq \emptyset$}
        \STATE $\text{improved} \leftarrow \text{False}$
        \STATE $u^* \leftarrow \text{None}$; $\rho^* \leftarrow 0$
        \FOR{$w$ \textbf{in} $S$}
            \STATE $\rho \leftarrow \text{Density}(C \cup \{w\})$
            \IF{$\rho \ge \gamma$ \AND $\rho > \rho^*$}
                \STATE $u^* \leftarrow w$; $\rho^* \leftarrow \rho$; $\text{improved} \leftarrow \text{True}$
            \ENDIF
        \ENDFOR
        \IF{$\text{improved}$}
            \STATE $C \leftarrow C \cup \{u^*\}$; $S \leftarrow S \setminus \{u^*\}$
        \ENDIF
    \ENDWHILE
    \IF{$|C| \ge s$ \AND $|C| > |C_{\mathrm{best}}|$}
        \STATE $C_{\mathrm{best}} \leftarrow C$
    \ENDIF
\ENDFOR
\RETURN $C_{\mathrm{best}}$
\end{algorithmic}
\end{algorithm}

\section{Appendix: Toy Logic Task Details}
\label{appendix:logic}
\subsection{Task Specification and Dataset}
The toy logic task is designed as a controlled synthetic setting where the target computation is perfectly specified by a known Boolean expression. The model is presented with a sequence of six input tokens, $t_0, \dots, t_5$, drawn uniformly from a predefined vocabulary. The target label is defined by the high-level causal model $\mathcal{H}$, which computes the expression $o_5 = ((t_2 \neq t_4) \land (t_0 \neq t_5)) \lor (t_1 = t_3)$.

For our analysis, we explicitly define the intermediate causal variables as follows:
\begin{align*}
    o_1 &:= (t_2 \neq t_4) \\
    o_2 &:= (t_0 \neq t_5) \\
    o_3 &:= (t_1 = t_3) \\
    o_4 &:= o_1 \land o_2 \\
    o_5 &:= o_4 \lor o_3
\end{align*}

\paragraph{Dataset Construction}
To format the inputs for the language model, we use an in-context learning template. Each prompt consists of 5 randomly sampled context examples (with their corresponding ground-truth Boolean labels) to establish the task format, followed by the target six-token sequence formatted as ``\texttt{t0,t1,t2,t3,t4,t5=}''. The dataset is constructed such that the intermediate variables (e.g., $o_3$) are balanced to be True with a probability of approximately 0.5. We train a 12-layer \texttt{GPT-2-small} ($\sim$117M parameters) on 2048 such examples, achieving 99.7\% task accuracy. 

Before any alignment search or graph construction, we rigorously filter the dataset to the $\mathcal{I}_{\mathrm{correct}}$ subset: we ensure that the model correctly predicts both the base input and the source input in isolation, verifying that any downstream failure under interchange intervention stems strictly from the alignment hypothesis, not from base model incompetence.

\subsection{Experimental Setup: Alignment Search}
\begin{figure}[htbp]
     \centering
     \begin{subfigure}[b]{0.45\textwidth}
         \centering
         \includegraphics[width=\textwidth]{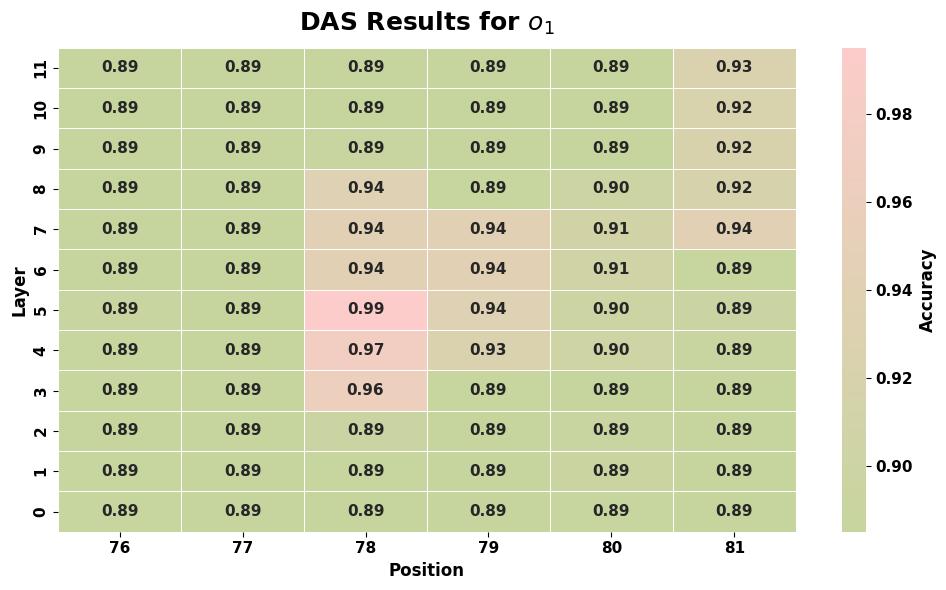}
     \end{subfigure}
     \hfill
     \begin{subfigure}[b]{0.45\textwidth}
         \centering
         \includegraphics[width=\textwidth]{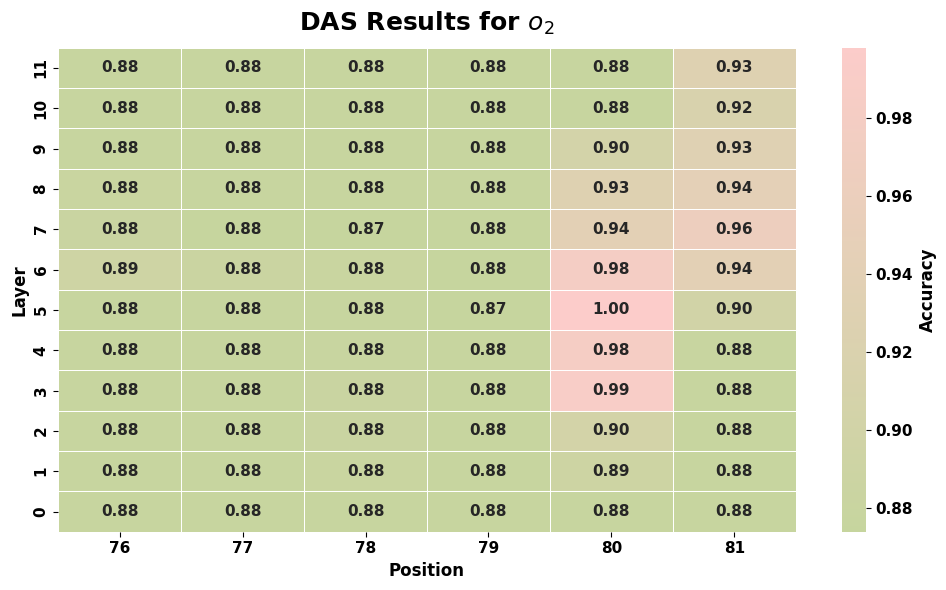}
     \end{subfigure}

     \vspace{0.5cm} 

     \begin{subfigure}[b]{0.45\textwidth}
         \centering
         \includegraphics[width=\textwidth]{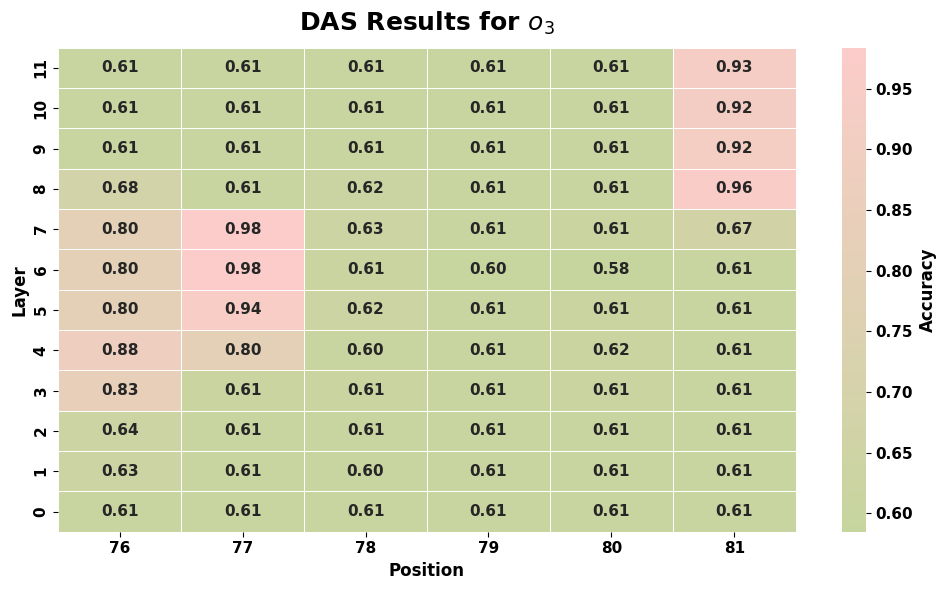}
     \end{subfigure}
     \hfill
     \begin{subfigure}[b]{0.45\textwidth}
         \centering
         \includegraphics[width=\textwidth]{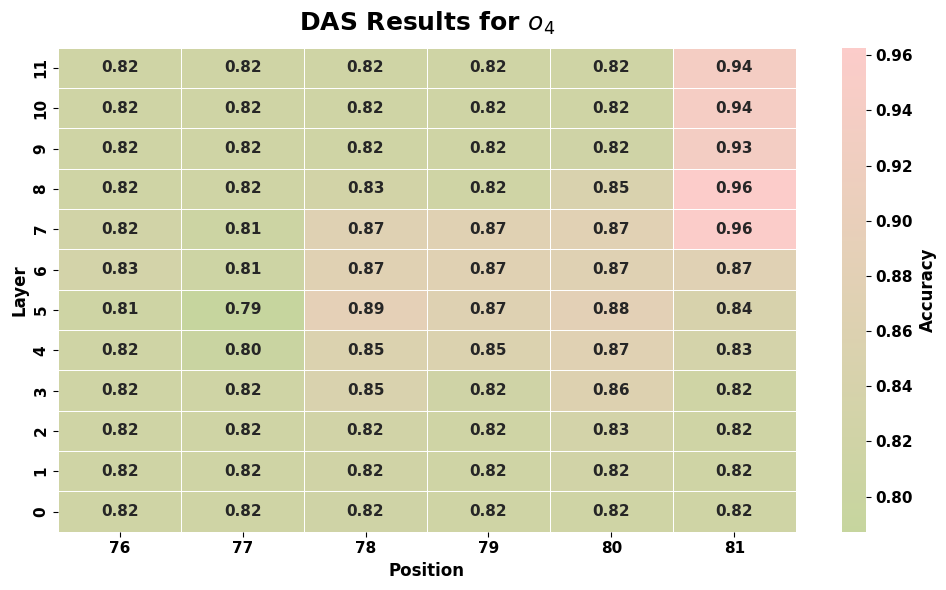}
     \end{subfigure}
        
     \caption{DAS Alignment Heat Maps for $o_1,o_2,o_3,o_4$. }
     \label{fig:logic_das}
\end{figure}

We use DAS to map the high-level variables ($o_4$, $o_5$) to the model's internal representations. We utilize the \texttt{pyvene} library to train DAS. The results are shown in \cref{fig:logic_das}. 

\paragraph{Training Details}
Because the causal variables are binary, we train a 1-dimensional orthogonal rotation matrix (\texttt{subspace\_dimension=1}) operating on the block output. For each candidate layer and token position, we train the DAS intervention for 5 epochs using the Adam optimizer (learning rate = $0.001$) with a batch size of 32. The objective minimizes the Cross-Entropy loss between the model's counterfactual output and the target causal model's predicted counterfactual label.

\subsection{Diagnostic Graph Construction}
Once the optimal candidate layer and position for an intermediate variable are identified via DAS, we construct the Interchangeability Graph to diagnose the alignment's structural faithfulness. For $N$ sampled inputs from the filtered dataset, we perform exhaustive directed interventions for all pairs $(i, j)$ using the trained DAS weight. An undirected edge is drawn between node $i$ and node $j$ if and only if the interchange intervention succeeds bidirectionally---meaning the model accurately predicts the counterfactual output when patching from $i \rightarrow j$, and independently when patching from $j \rightarrow i$. 

\paragraph{Partitioning}
We partition the resulting graph using a heuristic multi-seed greedy quasi-clique search (\cref{alg:multi_seed_quasi}). To accommodate slight representational noise inherent to neural activations, we apply a density threshold of $\gamma = 0.98$ and a minimum clique size of 2. The resulting clusters explicitly separate the input space into perfectly-interpreted (interchange-consistent) regions and under-interpreted regions.

\subsection{SAE-Based Generalization and Classification}
To determine if the boundaries of the perfectly interpreted subspaces are explicitly encoded in the model's feature space, we train classifiers using SAE features to predict bucket membership.

\paragraph{Feature Extraction and Classification}
For the nodes in the intervention graph, we extract the residual-stream activations at the targeted layer and token position. We encode these activations using pretrained SAEs for GPT-2 Small (specifically, the \texttt{gpt2-small-res-jb} release from \texttt{sae\_lens}). Using the resulting sparse feature activations, we train an $\ell_1$-regularized Logistic Regression classifier on an 80/20 train/test split. The model is trained to classify whether an input belongs to the perfectly-interpreted bucket or the other bucket. The high test accuracy of this classifier and the sparse set of non-zero $\ell_1$ coefficients allow us to identify the specific features (such as $o_4$) that dictate whether the high-level causal hypothesis holds.

\section{Entity Binding Task Details}

\subsection{Task Specification and Dataset}
We instantiate the entity binding task using the \texttt{filling\_liquids} task family. In this setting, the model is provided with a prompt containing 10 distinct entity groups. Each group follows a fixed template: \textit{"[Person] fills a [Container] with [Liquid]."} The prompt concludes with a query such as \textit{"Who filled the [Container]?"}, requiring the model to retrieve the specific [Person] associated with that container from the preceding context.

\paragraph{Entity Pools} To ensure high diversity and minimize accidental overlaps, we utilize the following entity pools:
\begin{itemize}
    \item \textbf{Persons:} John, Mary, Bob, Sue, Tim, Kate, Dan, Lily, Max, Eva, Sam, Zoe, Leo, Mia, Noah, Ava, Ben, Liz, Tom, Joy.
    \item \textbf{Containers:} cup, glass, bottle, mug, jar, pitcher, bowl, flask, tumbler, chalice, vessel, container, tank, can, tube, vial, goblet, stein, carafe, decanter.
    \item \textbf{Liquids:} beer, wine, water, juice, milk, coffee, tea, soda, lemonade, smoothie, soup, broth, sauce, syrup, oil, honey, cider, nectar, punch, tonic.
\end{itemize}

\paragraph{Dataset Construction} We generate a dataset of 1,024 samples. Following our diagnostic pipeline, we filter this dataset to include only instances where \texttt{Gemma-2-2B-Instruct} correctly predicts the target entity for both the base and counterfactual inputs. After filtering, the dataset is split into training (80\%) and testing (20\%) sets.

\subsection{Experimental Setup}
We use \texttt{pyvene} to conduct vanilla interchange interventions. We test the \textit{positional hypothesis}, which posits that the model's retrieval mechanism is mediated by a high-level variable $q_{\mathrm{group}}$ representing the context position of the queried group. 

\paragraph{Alignment Procedure} Counterfactuals are generated by swapping the positions of entity groups while maintaining the templatic structure. We perform full-vector patching on the residual stream  at the final query token position across all layers. A faithful abstraction requires that patching the internal representation from a source input into a base input causes the model to retrieve the entity corresponding to the source's queried group position.

\subsection{Diagnostic Results}
\begin{figure}[!htb]
    \centering
    \includegraphics[width=0.7\linewidth]{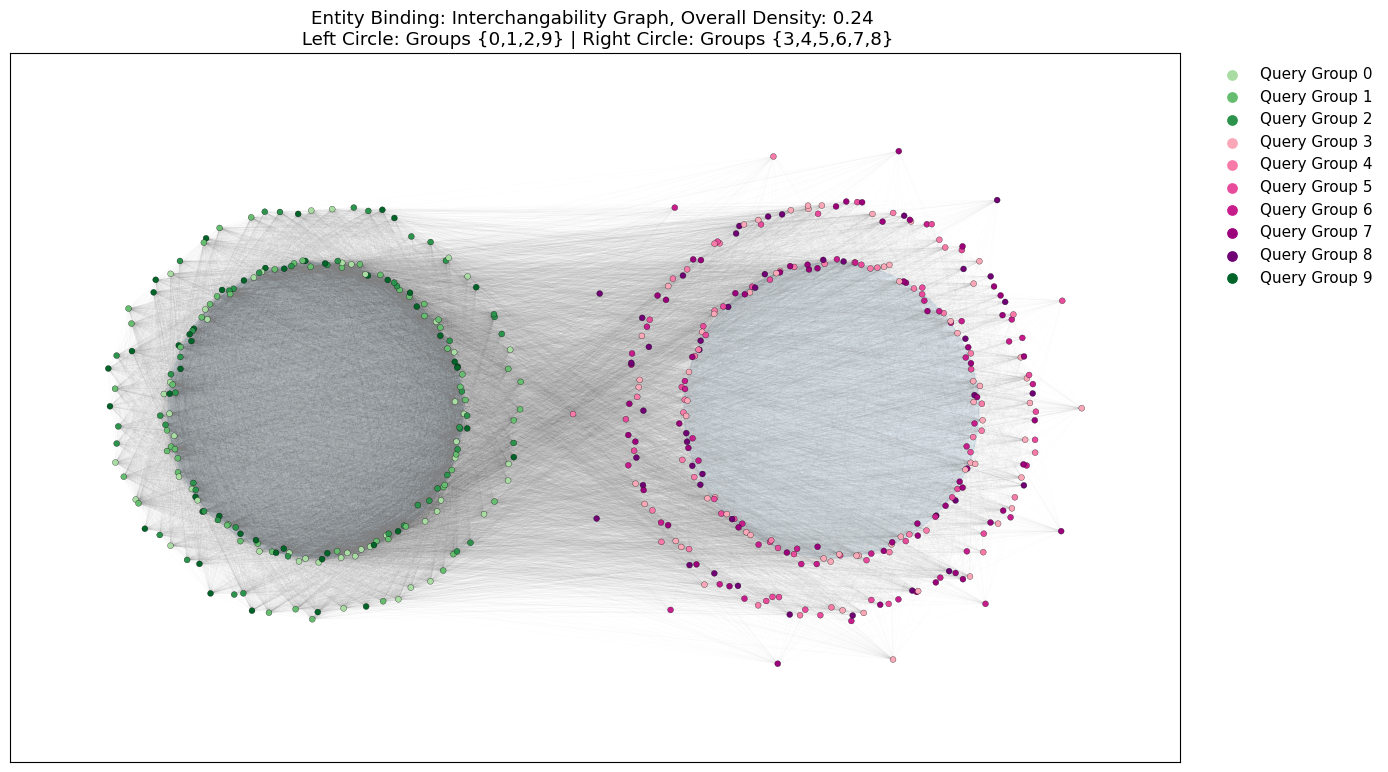}
    \caption{\textbf{Interchangeability Graph for Entity Binding.} The graph nodes represent individual inputs, and edges denote perfect pairwise interchangeability. The emergent community structure corresponds to the "Target Buckets" where the positional hypothesis is perfectly faithful, primarily at the start and end of the prompt sequence.}
    \label{fig:InterchangeabilityGraph}
\end{figure}

We sample $512$ correctly answered prompts from the \texttt{filling\_liquids} input space, perform pairwise interchange interventions, and construct an Interchangeability Graph whose vertices represent individual inputs. We analyze the resulting IIA to construct the interchangeability graph shown in \cref{fig:InterchangeabilityGraph}. 

For the classification step of our diagnosis, we utilize internal features from \texttt{Gemma Scope} \cite{lieberum2024gemmascope} to determine if the boundaries between well-interpreted and under-interpreted regions are explicitly encoded in the model's feature space.

\section{Entangled Factual Recall Task Details}

\subsection{Task Specification and Dataset}
We evaluate the entangled factual recall setting using the RAVEL benchmark \citep{huang2024ravel} on the \texttt{Llama-3.1-8B} model. We focus on the \textsc{Language} attribute of the \textit{city} entity type. The input space consists of prompts where the base and source inputs share the exact same template but differ only in the city name (e.g., \textit{"People in San Francisco speak..."} vs. \textit{"People in Paris speak..."}). 

To ensure our interventions target the specific entity representation, we extract the residual stream activations at the final token of the city name. The dataset is filtered to strictly include instances where the base model correctly predicts the factual target. The filtered dataset is then split into an 80\% training set and a 20\% testing set.

\subsection{Experimental Setup: MDAS}
We employ Multi-task Distributed Alignment Search (MDAS) to learn a disentangled subspace that isolates the \textsc{Language} attribute from other co-encoded attributes (\textsc{Continent}, \textsc{Country}, \textsc{Latitude}, \textsc{Longitude}, and \textsc{Timezone}). 

\paragraph{MADS Loss}  Following \cite{huang2024ravel}, given an entity $E$ and an attribute $A$ with a ground-truth value $A_E$ (e.g., \textsc{Paris} and \textsc{Continent}), we seek to learn a feature $F_A$. A high \textbf{Cause} score indicates that intervening on this representation successfully transfers a source attribute value $A_{E'}$ from a counterfactual input $x'$ to the model's prediction for a base input $x$, optimized as:
\begin{equation}\label{eq:loss_cause}
\mathcal{L}_{Cause}(A, F_A, \mathcal{M}) = \text{CE}(\text{II}(\mathcal{M}, F_A, x, x'), A_{E'})
\end{equation}
Conversely, a high \textbf{Iso} (isolation) score ensures the intervention is surgical and does not alter other attributes $A^* \in \mathcal{A} \setminus \{A\}$. For a prompt $x^*$ querying a different attribute $A^*$ with ground-truth value $A_E^*$ (e.g., \textsc{Language}), the isolation loss ensures the model still predicts the original value $A_E^*$ despite the intervention on $F_A$:
\begin{equation}\label{eq:loss_iso}
\mathcal{L}_{Iso}(A, F_A, \mathcal{M}) = \frac{1}{|\mathcal{A} \setminus \{A\}|} \sum_{A^* \in \mathcal{A} \setminus \{A\}} \text{CE}(\text{II}(\mathcal{M}(x^*), F_A, x'), A_E^*)
\end{equation}
The MDAS training  loss is the summation of $\mathcal{L}_{Cause}$ and $\mathcal{L}_{Iso}$.  

\paragraph{Training Details} Rather than exhaustively training across all layers, we first perform a coarse search using a large layer gap across subspace dimensions $k \in \{32, 128, 512, 2048\}$. Based on the preliminary alignment results from this initial sweep, we zoom in on the most promising regions to localize the optimal subspace and identify layer 14 as the ideal intervention site. To optimize for both causal transfer and isolation, we construct a weighted training distribution: pairs evaluating the target attribute (\textsc{Language}) are sampled at a 5:1 ratio compared to pairs evaluating the other five off-target attributes. This forces the subspace to maximize the interchange intervention accuracy (IIA) for \textsc{Language} while ensuring interventions do not alter the model's predictions for queries about a city's continent or timezone.

\begin{figure}[!htb]
    \centering
    \includegraphics[width=0.5\linewidth]{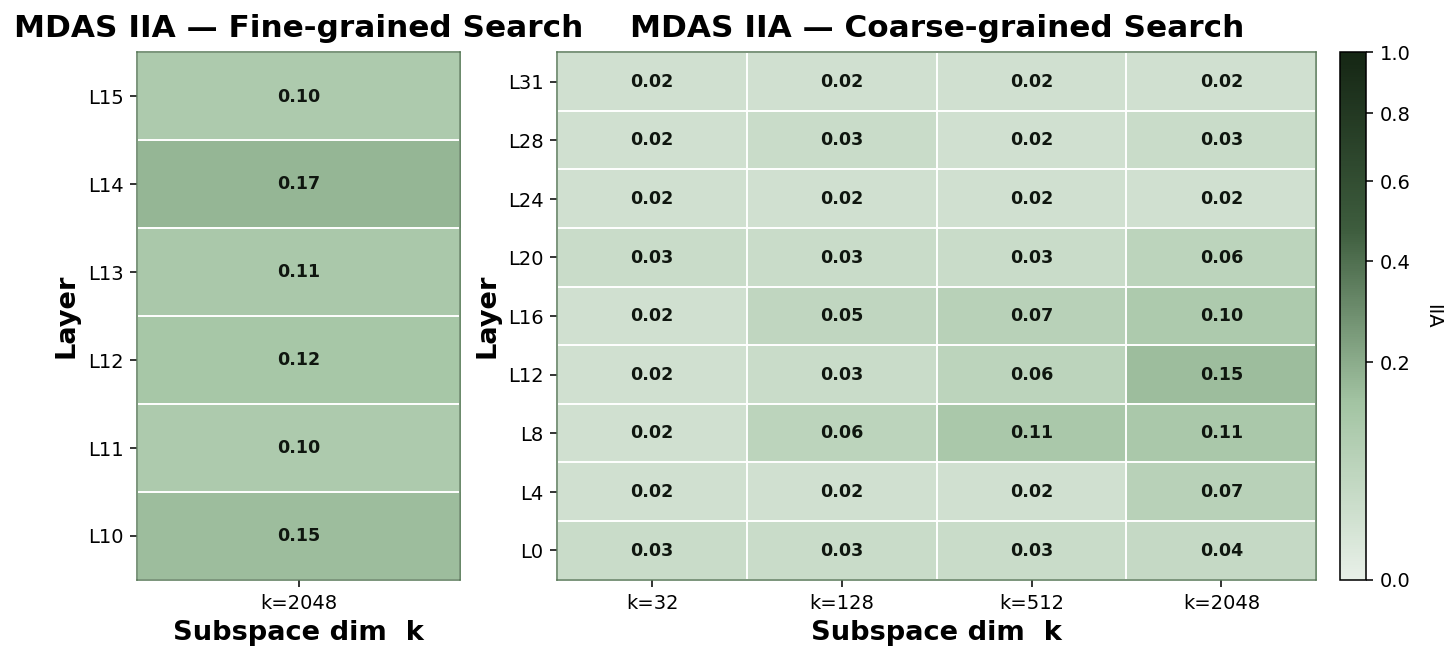}
    \caption{MDAS Alignment Results for Entangled Factual Recall Task}
    \label{fig:mdas}
\end{figure}

\subsection{Diagnostic Results and Classification}
Following the MDAS optimization, we apply our diagnostic pipeline to the test set. 

\paragraph{Graph Construction and Partition} We perform exhaustive pairwise interchange interventions. An undirected edge is formed between two inputs if and only if the bidirectional interchange intervention (source $\rightarrow$ base, and base $\rightarrow$ source) is perfectly consistent with the high-level causal model. We partition this graph using \cref{alg:multi_seed_quasi} with a density threshold of $\gamma = 0.98$ and a minimum clique size of 2, which successfully isolates the language-specific clusters (English, Spanish) discussed in the main text. We find that with fewer buckets, the partition first separates English from non-English cities, and then Spanish splits off as a second clean group.

\paragraph{SAE-Based Generalization} To test if these boundaries are encoded in the model's internal feature space, we train a classifier using sparse features from \emph{Llama Scope} \citep{he2024llamascope}. We extract the residual stream activations at layer 14 for the graph nodes and encode them through the corresponding Llama SAE. 

Using the resulting sparse feature activations, we train a multi-class $\ell_1$-regularized Logistic Regression classifier to predict the quasi-clique bucket labels. To interpret the structural differences between buckets, we extract the highest-weight SAE features by analyzing the $\ell_1$ coefficients.

\end{document}